\documentclass[conference]{IEEEtran}
\IEEEoverridecommandlockouts
\usepackage{cite}
\usepackage{amsmath,amssymb,amsfonts}
\usepackage{algorithmic}
\usepackage{graphicx}
\usepackage{tabularx}
\usepackage{textcomp}
\usepackage{xcolor}
\usepackage{booktabs}
\usepackage{float}
\usepackage{subcaption}
\usepackage{multirow}

\def\BibTeX{{\rm B\kern-.05em{\sc i\kern-.025em b}\kern-.08em
    T\kern-.1667em\lower.7ex\hbox{E}\kern-.125emX}}

\PassOptionsToPackage{hyphens}{url}    
\usepackage{tikz,xcolor,hyperref}
\usepackage{microtype}
\usetikzlibrary{shapes,arrows,positioning}

\definecolor{lime}{HTML}{A6CE39}
\definecolor{prpa}{HTML}{648fff}
\definecolor{prpb}{HTML}{785ef0}
\definecolor{prpc}{HTML}{dc267f}
\definecolor{prpd}{HTML}{fe6100}
\definecolor{prpe}{HTML}{ffb000}

\DeclareRobustCommand{\orcidicon}{
	\begin{tikzpicture}
	\draw[lime, fill=lime] (0,0) 
	circle [radius=0.16] 
	node[white] {{\fontfamily{qag}\selectfont \tiny ID}};
	\draw[white, fill=white] (-0.0625,0.095) 
	circle [radius=0.007];
	\end{tikzpicture}
	\hspace{-2mm}
}
\foreach \x in {A, ..., Z}{%
	\expandafter\xdef\csname orcid\x\endcsname{\noexpand\href{https://orcid.org/\csname orcidauthor\x\endcsname}{\noexpand\orcidicon}}
}

\usepackage{todonotes}    

\begin{document}

\title{Image-Based Leopard Seal Recognition:
Approaches and Challenges in Current Automated Systems}

\author{\IEEEauthorblockN{Jorge Yero Salazar\orcidA{}, Pablo Rivas\orcidC{}, \emph{Senior, IEEE}}
\IEEEauthorblockA{School of Engineering \& Computer Science\\
Department of Computer Science\\
Baylor University, Texas, USA\\
Email: \{Jorge\_Yero1, Pablo\_Rivas\}@Baylor.edu}
\and
\IEEEauthorblockN{Renato Borras-Chavez\orcidD{}, Sarah Kienle\orcidE{}}
\IEEEauthorblockA{College of Arts \& Sciences\\
Department of Biology\\
Baylor University, Texas, USA\\
Email: \{Renato\_Borras-Chavez, Sarah\_Kienle\}@Baylor.edu}
}

\maketitle

\begin{abstract}
This paper examines the challenges and advancements in recognizing seals within their natural habitats using conventional photography, underscored by the emergence of machine learning technologies. We used the leopard seal, \emph{Hydrurga leptonyx}, a key species within Antarctic ecosystems, to review the different available methods found. As apex predators, Leopard seals are characterized by their significant ecological role and elusive nature so studying them is crucial to understand the health of their ecosystem. Traditional methods of monitoring seal species are often constrained by the labor-intensive and time-consuming processes required for collecting data, compounded by the limited insights these methods provide. The advent of machine learning, particularly through the application of vision transformers, heralds a new era of efficiency and precision in species monitoring. By leveraging state-of-the-art approaches in detection, segmentation, and recognition within digital imaging, this paper presents a synthesis of the current landscape, highlighting both the cutting-edge methodologies and the predominant challenges faced in accurately identifying seals through photographic data.
\end{abstract}

\begin{IEEEkeywords}
Vision Transformer, Object Detection, Semantic Segmentation, Leopard Seals, Seals Detection, Computer Vision
\end{IEEEkeywords}

\section{Introduction}
\label{sec:intro}

Like other apex predators, the leopard seal, \emph{Hydrurga leptonyx}, is a species that plays a crucial role in Antarctic ecosystems and has captivated researchers with its distinctive ecological functions and behaviors. Nonetheless, researching these hard-to-find animals in their remote habitats is challenging. Conventional methods of monitoring and gathering data are labor-intensive, time-consuming, and tend to provide restricted information because of the leopard seal's elusive nature. The advent of machine learning has provided a breakthrough in overcoming these limitations. By integrating machine learning techniques, wildlife research has embraced automated recognition methods that significantly improve the efficiency and accuracy of species monitoring. 

Building upon this technological advancement, projects focusing on the recognition of seals have predominantly utilized conventional photographs as the primary data source. We characterize these conventional photographs based on three criteria: they are composed of the standard Red-Green-Blue three-channel color model; their resolution is typical of images taken with handheld cameras ; and the focal subject, such as a seal, occupies at least 20\% of the image area. This categorization generally encompasses images captured by handheld devices and some drone-operated cameras, excluding satellite imagery. Satellite photographs are typically omitted due to their incorporation of additional spectral channels, ultra-high resolution, and a relatively small representation of objects of interest, necessitating distinct analytical approaches~\cite{pritt2017satellite}. Within the scope of digital image recognition, accurately identifying the location of seals hinges on the critical processes of detection and segmentation.

Developing any project centered on recognition, along with the essential detection and segmentation stages, fundamentally begins with an examination of the existing state of the art (SOTA) of this area. Therefore, this paper presents a detailed synthesis of the latest advancements in seal recognition through conventional photographs, elucidating the most cutting-edge models and methodologies in the field and highlighting the principal challenges and limitations inherent to the aforementioned processes. To accomplish this, the subsequent sections are organized and presented in a manner that addresses three fundamental questions: 

\begin{enumerate}
    \item What are the most successful SOTA approaches for detecting and segmenting seals using conventional photographs? How do these approaches compare, and what are the biggest challenges?
    \item Why have vision transformers(ViT) gained popularity recently compared to other SOTA models in image recognition tasks, and what is their potential application for recognizing seals in surface marine environments?
    \item What are the primary challenges associated with recognizing seals from Computer Vision (CV) and Machine Learning (ML)? How do these challenges impact the development and performance of automated detection systems?
\end{enumerate}
We provide examples of some of the revised methodologies and approaches, using leopard seals as our main model and our own dataset of the species. 

\section{Approaches for detection and segmentation of seals using conventional photographs}

In the rapidly evolving field of Computer Vision, object detection and image segmentation are foundational tasks that enable machines to interpret and understand visual data accurately and in-depth. Object detection refers to identifying and locating objects within an image. This involves classifying the types of objects present and pinpointing their positions with bounding boxes or similar markers. Image segmentation takes this further by partitioning an image into multiple segments or pixels to simplify its representation or make it more meaningful. It is a critical step in distinguishing between the object of interest and the background or between different objects in an image.

This section will examine the latest advancements in object detection and image segmentation techniques for seals in conventional photographs. The section also explores image segmentation strategies used in previous studies.

\subsection{Seals detection}\label{sec:object_detection}

% We focus on two main elements. Firstly, our subject of interest is a specific type of animal: seals. Secondly, our analysis is centered on a particular medium: conventional photographs.

This revision focuses on studies incorporating object detection with pinnipeds (true and ear seals) as study models. We found that Pinniped´s studies often focus either on seals using non-traditional imagery, like satellite photos 
\cite{detection_seal_aerial_salberg2015detection,detection_seal_satellite_gonccalves2020sealnet,detection_seal_satellite_gonccalves2022sealnet2}, or on different species using conventional images~\cite{detection_primates_conventional_manual_deb2018face,detection_lemur_conventional_manual_crouse2017lemurfaceid,detection_cetacean_conventional_patton2023deep}, and rarely tackling both simultaneously. We found only one study that examines seals in conventional photographs and relies on manual face landmark detection~\cite{detection_seal_conventional_manual_birenbaum2022sealnet}.

We identified three studies using non/conventional photographs employing Convolutional Neural Networks (CNNs~\cite{cnn_lecun1989backpropagation}) for object detection~\cite{detection_seal_aerial_salberg2015detection,detection_seal_satellite_gonccalves2020sealnet,detection_seal_satellite_gonccalves2022sealnet2}. Alternatively, suppose we keep the domain of interest but allow different species as objectives. In that case, we analyze three studies that use manual detection methods and one study adopting both You Only Look Once (YOLO)~\cite{yolov1_redmon2016you} and Detector with image classes (Detic)~\cite{detic_zhou2022detecting} technologies. For a more precise overview of the research landscape and how it intersects with our interests, refer to Table \ref{tab:detection_list}, which organizes these studies based on the two critical aspects of our inquiry.

\begin{table}[h!]
    \centering
    \caption{Overview of Research Studies on Object Detection in Different Species and Using Different Photographic Domains}
    \begin{tabular}{c|l|l|l}
    \toprule
         \textbf{Paper} & \textbf{Study Object} & \textbf{Photography Domain} & \textbf{Methodology}  \\ \hline
        \cite{detection_seal_conventional_manual_birenbaum2022sealnet} & \textbf{seals} & \textbf{conventional} & Manual\\
        \cite{detection_primates_conventional_manual_deb2018face} & primates & \textbf{conventional} & Manual \\
        \cite{detection_lemur_conventional_manual_crouse2017lemurfaceid} & lemur & \textbf{conventional} & Manual\\
        \cite{detection_cetacean_conventional_patton2023deep} & cetacean & \textbf{conventional} & YOLO v5, Detic \\
        \cite{detection_seal_aerial_salberg2015detection} & \textbf{seals} & satellite & R-CNN \\
        \cite{detection_seal_satellite_gonccalves2020sealnet} & \textbf{seals} & satellite & U-Net \\
        \cite{detection_seal_satellite_gonccalves2022sealnet2} & \textbf{seals} & satellite & Fast R-CNN \\
    \bottomrule
    \end{tabular}
    \label{tab:detection_list}
\end{table}

While there are indeed studies focusing on seal detection in non-conventional images, we excluded these studies due to the significant differences in the domain images and the specific pre-processing that is required for them. The images used in these studies typically have very high resolution, resulting in the region containing the seal proportionally much smaller than what is found in conventional photos. Additionally, these non-conventional images often feature more than three color channels in contrast to conventional ones that contain only three. A prevalent method in handling such high-resolution images is to segment them into smaller patches before feeding them into the detection model. This patch-based approach is necessary to manage the large image sizes and the detailed information content, which is not directly applicable to the domain of conventional photography that our research is focused on. Thus, we focus on studies that align with our criteria of using conventional images as their primary domain with different species (ref. Table \ref{tab:detection_list}).

\subsubsection{Seal detection on conventional photographs}

As previously noted, the sole study identified that applies object detection to seals in conventional photographs is the work in~\cite{detection_seal_conventional_manual_birenbaum2022sealnet}. This research aims to establish a model capable of recognizing individual seals by their facial features. Although facial recognition does not directly correspond to the main objective of this section, the initial phase of their approach, which involves detecting seal faces, bears relevance.

The detection phase employs a CNN, details of which were not extensively described in the paper. The methodology includes downscaling the image to reduce dimensionality and augmenting the number of channels using a series of filters. This is followed by applying convolutional layers normalized in batches and utilizing the ReLU activation function.

The study in question underscores the significance of eye landmarks in orienting seal face images for consistent recognition. It's important to highlight that accurate facial recognition, particularly in humans, has been shown to rely heavily on the positioning of the eyes~\cite{face_recognition_jain2011handbook}.

This investigation successfully culminates in a model that automates the detection of seal faces, marking a significant stride toward individual seal recognition. However, manual intervention is required during data preparation. Following the initial automatic detection of faces, researchers must manually pinpoint the eyes, nose, and mouth locations. The resultant coordinates are critical for the software to automatically perform subsequent image alignment and cropping tasks.

\subsubsection{Other species detection on conventional photographs}

The methodology proposed in~\cite{detection_seal_conventional_manual_birenbaum2022sealnet} is similar in other studies~\cite{detection_lemur_conventional_manual_crouse2017lemurfaceid},~\cite{detection_primates_conventional_manual_deb2018face} that maintain the conventional photography domain but focus their attention on different species, such as lemurs and primates respectively. The primary goal of these works was to identify individuals through facial recognition, in which face detection is also necessary as part of the process. Unlike the seal face detection approach~\cite{detection_seal_conventional_manual_birenbaum2022sealnet}, these studies do not employ automatic face detection methods but rely on manually annotated landmarks for the face location. Furthermore, estimating the face's width and height is based on the inter-pupil distance, which is the distance between the centers of the two eyes. This measurement is then used to scale the image so that the eyes are positioned at specific proportions relative to the edges of the cropped face image. Both studies implement a face alignment as a pre-processing step similar to the previously mentioned work.

Furthermore, using conventional photographs from cetaceans, studies have implemented automatic full-body object detection~\cite{detection_cetacean_conventional_patton2023deep}. This research introduces a multi-species individual identification model leveraging two ArcFace heads, a cutting-edge technique in human facial recognition~\cite{arcface_deng2019arcface},~\cite{arcface_deng2020sub}. These heads perform dual functions, classifying species and individual identities, thus allowing species to share information and parameters within the network.

The study employs an object detection phase pertinent to our research question as a component of the identification process. This phase aims to crop the image, effectively reducing background noise that may hinder the identification-matching process. The researchers integrated four distinct detectors, each activated according to a specific probability—three based on YOLOv5~\cite{yolov1_redmon2016you} and one based on Detic~\cite{detic_zhou2022detecting}.

\subsubsection{Comparing approaches and identifying challenges}\label{sec:object_detection_discussion}

The comparison of state-of-the-art (SOTA) approaches for detecting seals in conventional photographs reveals a landscape where very specific methodologies are applied, each tailored to the unique challenges presented by the subject matter and the medium. The primary difference among the approaches is the balance between manual and automated detection methods. The manual face landmark detection approach suffers from scalability issues due to its manual nature. In contrast, automatic full-body detection offers a more scalable and less labor-intensive solution.

In addressing our specific objective of detecting seals in conventional photographs, the task simplifies to a binary classification problem within a bounding box, where the model must discern whether the box contains a seal or not. This framing does not align with the core strength of Detic, which is designed to manage a wide array of classes for bounding boxes and to identify objects outside the predefined vocabulary. Consequently, our focused application may only partially utilize Detic's ability to handle extensive class variety and predict out-of-vocabulary objects. Conversely, YOLOv5 is celebrated for its optimal balance between speed and accuracy, making it an ideal candidate for applications that demand rapid processing without significantly compromising precision. This attribute of YOLOv5 could be particularly beneficial for scenarios requiring real-time monitoring or immediate analysis.

Seal detection on conventional photographs has many challenges, but the biggest one is the lack of previous work that does not rely on manual annotations. Also, conventional photographs vary widely in quality and resolution, posing significant challenges in identifying small or subtle features crucial for species identification. Manual methods, while precise, are not scalable and require substantial human effort. Automated methods must be adaptable to different species and contexts, which can be challenging given the variability in animal appearances and environmental conditions. Furthermore, high-quality, annotated datasets are crucial for training effective models, yet such datasets are often scarce or difficult to compile for wildlife studies.

The YOLO framework, noted for its agility and accuracy, emerges as a promising solution to these constraints. However, YOLO's true efficacy and applicability over manual and automated approaches like Detic can only be accurately gauged through direct application and comparative analysis. Such empirical testing is crucial to validate YOLO's potential advantages in speed and scalability and explore its adaptability to the diverse quality and resolution of conventional photographs, which pose challenges in accurately identifying seals.
In conclusion, while manual and automated detection methods offer valuable insights into seal detection in conventional photographs, their efficacy and applicability vary based on the research objectives, available resources, and specific challenges of working with wildlife imagery.

\subsection{Seals segmentation}

In the context of image segmentation, two primary tasks exist: semantic segmentation and instance segmentation. Although both tasks can differentiate between pixels that represent a seal and those that do not in an image, our focus will be on instance segmentation. This choice is motivated by its ability to identify and segment individual instances of the same object separately. In contrast, semantic segmentation treats multiple objects of the same category as a single, unified segment if they are in close proximity or touching. This distinction makes instance segmentation particularly valuable for applications requiring precise identification and differentiation of each object, regardless of its category similarity to others in the image.

\subsubsection{Seal segmentation on conventional photographs}

\begin{figure}[b!]
    \centering
    \includegraphics[width=0.7\linewidth]{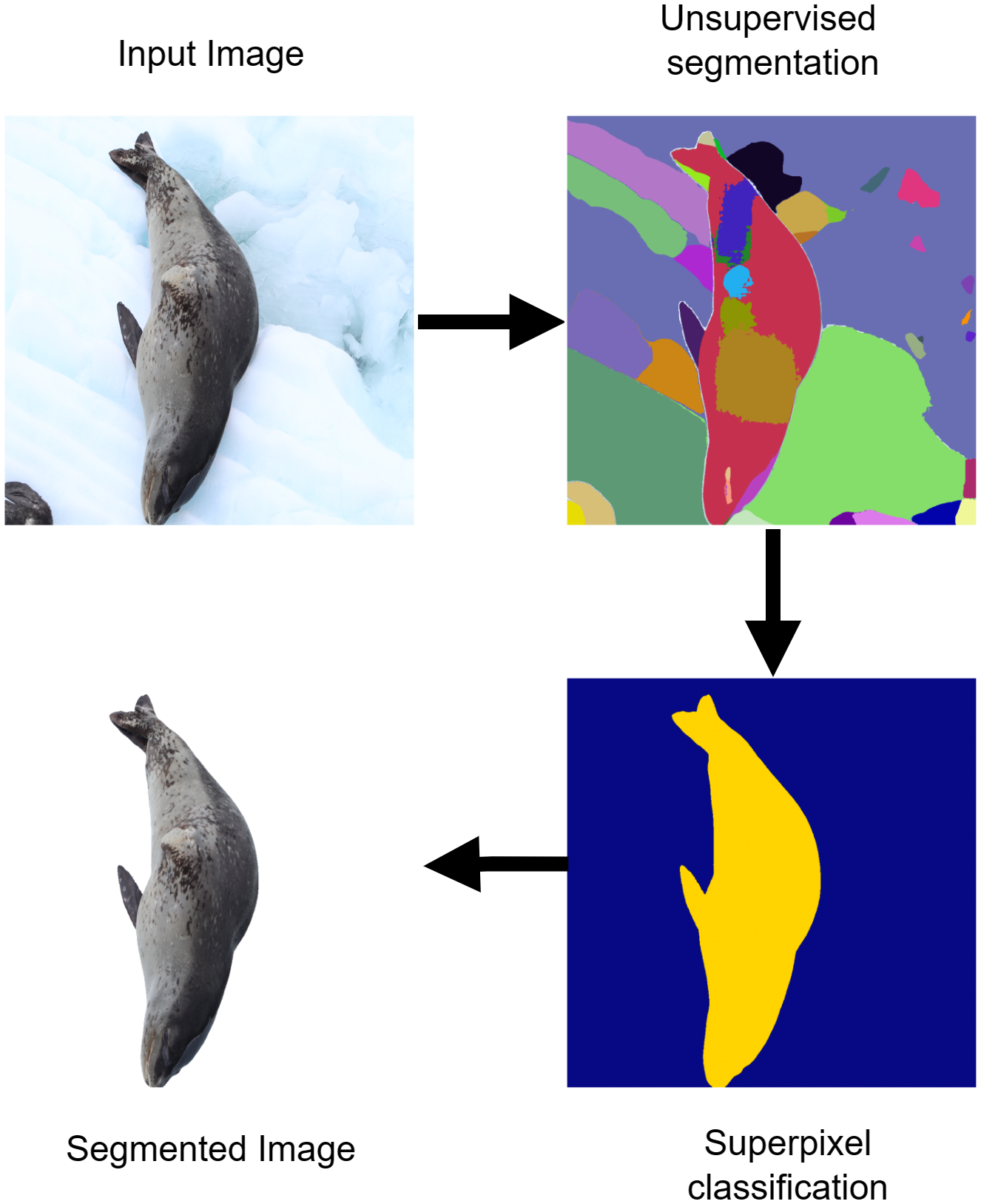}
    \caption{Methodology for segmentation following the work in~\cite{seal_segmentation_zhelezniakov2015segmentation}. The image was created based on their image methodology but kept only the part aligned with our question.}
    \label{fig:seal_segmentation}
\end{figure}

A single study emerges from reviewing methodologies for seal segmentation ~\cite{seal_segmentation_zhelezniakov2015segmentation} emerges as a primary reference. This research explicitly targets the accurate identification of Saimaa ringed seals (\textit{Pusa hispida)} . While the focus on individual seal recognition does not directly correlate with our scope, the methodology proposed in~\cite{seal_segmentation_zhelezniakov2015segmentation} is noteworthy. Their approach involves an initial segmentation phase to isolate the seal from the background, followed by applying a separate recognition algorithm to the segmented image. Fig. \ref{fig:seal_segmentation}  uses our own dataset and provides an enhanced visualization of the methodology described in~\cite{seal_segmentation_zhelezniakov2015segmentation}.

The study highlights the utility of segmentation, mainly because all images are captured during the same seasonal period each year with stationary cameras. This consistency in imaging conditions allows the algorithm to potentially rely on highly similar backgrounds for identification when analyzing the entire image or specific bounding boxes rather than the seals themselves. Zhelezniakov et al.~\cite{seal_segmentation_zhelezniakov2015segmentation} prefer semantic segmentation rather than instance segmentation based on the assumption that each input image contains only a single seal. Given this specific constraint, the task can effectively be treated as instance segmentation since each image contains only one instance for segmentation.

As previously stated, the segmentation process for identifying Saimaa ringed seals in images employs a sophisticated methodology that begins with unsupervised segmentation to divide the image into superpixels or minor, coherent pixel clusters. This initial step utilizes a state-of-the-art algorithm combining a Globalized Probability of Boundary (gPb) detector, Oriented Watershed Transform (OWT), and Ultrametric Contour Map (UCM), which together generate superpixels by thresholding a weighted contour image~\cite{unsupervised_semantic_segmentation_arbelaez2009contours}~\cite{unsupervised_semantic_segmentation_arbelaez2010contour}. Adjusting the threshold value, the size of the superpixels can be controlled. This adjustment helps pick sufficiently large superpixels for reliable classification while preventing the seal and background from being combined into a single superpixel.

Following unsupervised segmentation, texture features are extracted from the superpixels using Blur Invariant Local Phase Quantization (LPQ) descriptors~\cite{ojansivu2008blur}. These features are particularly suited for processing camera trap images, which often exhibit blur and significant variations in illumination, as they are based on phase information that is invariant to such distortions.

The final step involves classifying each superpixel as either seal or background using a Support Vector Machine (SVM) classifier trained on manually labeled superpixels. This supervised classification process effectively distinguishes seal segments from the background, with the combined seal-classified superpixels forming the final segmented image of the seal for further identification processes. The classification accuracy is determined by intersection over union (IoU) over a certain threshold. This ensures that only those segments with significant conformity to the actual shape and location of the seal are considered accurately classified, thereby maintaining high precision in the segmentation process.

\subsubsection{Segment Anything Model}

Two studies were identified that discuss segmentation methods for images of different species (see~\cite{segmentation_animals_water_pollicelli2020roi,segmentation_animalli2021marine}). Although these investigations enrich the theoretical foundation of image segmentation, they diverge from our research objectives. The first study concentrates on images of animals swimming at the water's surface, while the latter uses underwater imagery as its domain. 

\begin{figure}[t!]
\centering
  \centering
  \includegraphics[width=\linewidth]{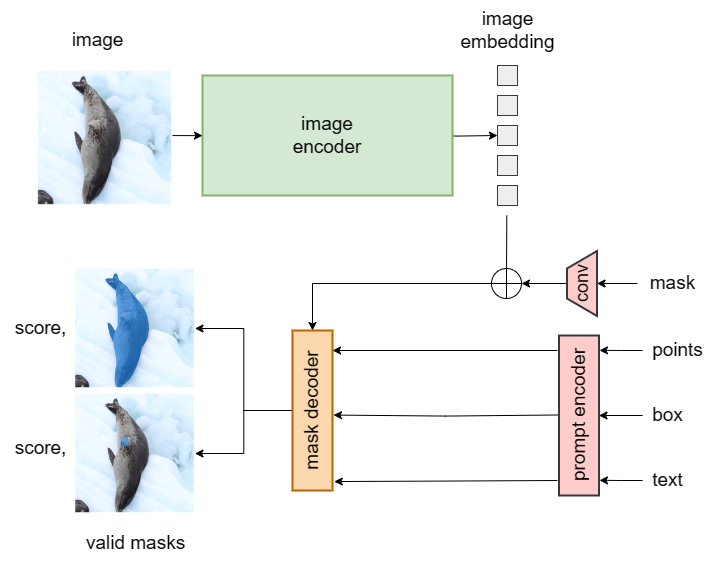}
  \caption{Segment Anything Model (SAM) overview. Image was based on the architecture figure from~\cite{segment_anything_kirillov2023segment} using our database.}
  \label{fig:segment_anything_overview}
\end{figure}

Consequently, our focus shifts towards one research encompassing a broader range of domains, including conventional imagery. The research conducted by~\cite{segment_anything_kirillov2023segment} has established a foundational model in segmentation, named Segment Anything Model (SAM), earning widespread recognition within the scientific community as evidenced by its impressive tally of 2,298 citations within the first eleven months post-publication. Although this research is not directly related to seal segmentation, the innovative approach of the model suggests that it can be adapted to various tasks through zero-shot transfer using prompt engineering. This advantage enables the application of the model for seal segmentation without the need for fine-tuning or manual annotation of seal segments, thereby demonstrating its adaptability and potential for expansive applications across different domains (Fig. \ref{fig:segment_anything_overview}).

The SAM architecture, illustrated in Fig. \ref{fig:segment_anything_overview}, processes an input image through an image encoder to generate an embedding. This embedding is then used to produce object masks responding to input prompts. The encoder's core is a Mask Auto Encoder (MAE), leveraging a pre-trained ViT for image encoding~\cite{mae_he2022masked, vit_dosovitskiy2020image}. Notably, the approach to querying with text diverges from other methods. It serves primarily as a conceptual demonstration for converting text to masks, though the models for this specific functionality have not been made public.

The image encoder remains consistent for other types of queries—points, boxes, and masks. Points and box queries are translated into positional encodings. Mask embeddings are generated through a CNN encoder added to the image embedding. The model outputs a list of pairs, each consisting of a mask and the associated confidence level for that mask.

The model is noteworthy for its versatility and efficiency in segmenting a wide range of objects within images, even those not pre-defined or labeled in existing datasets. This is a significant advancement because traditional segmentation models often require extensive training on large, labeled datasets that cover all possible objects and scenarios they might encounter. Such requirements make them less adaptable to new or rare objects. SAM aims to overcome these limitations by using an approach that can generalize from known objects to novel ones, allowing it to perform segmentation tasks on a broader array of images without needing specific training for each new object type.

Further evidence of the model's superiority over existing research is its outstanding performance in interactive segmentation, surpassing state-of-the-art models on 23 distinct datasets~\cite{segment_anything_kirillov2023segment}. Superior outcomes in human evaluation underscore this achievement and mean intersection over union (mIoU), particularly noting the model's accuracy in segmenting an object given its central point. Additionally, the model demonstrates comparable performance to the advanced ViTDet-H~\cite{vitdec_li2022exploring} model across various datasets, including COCO~\cite{coco_lin2014microsoft} and LVIS v1~\cite{lvis_gupta2019lvis}, in terms of average precision (AP). This is notable given that ViTDet-H was specifically trained on these datasets, whereas the Segmenting Anything Model was not. Additionally, in the realm of instance segmentation, where SAM was prompted with the bounding box of the target object, it not only matched but, based on human evaluations, exceeded the performance of ViTDet-H. This suggests that while ViTDet-H may rely on biases inherent in the COCO and LVIS datasets, the SAM approach, free from such training data biases, offers a more generalized and accurate segmentation capability.

\subsubsection{Comparing approaches and identifying challenges}

The superior capabilities of the Segment Anything Model are evident not only because of its innovative approach to image segmentation but also because of its ability to address issues that other studies in this field have yet to be able to overcome. For instance, the study of ~\cite{seal_segmentation_zhelezniakov2015segmentation} relies heavily on manual intervention, requiring extensive annotations for both identifying the seals (ground truth) and classifying superpixels generated by an unsupervised segmentation algorithm. This manual classification process is not only time-consuming but also susceptible to errors. Moreover, the reliance on texture feature extraction can lead to inconsistencies influenced by varying weather conditions and background elements. The Segment Anything Model addresses these limitations using zero-shot instance segmentation, which eliminates the need for detailed image segment annotations and requires only the provision of bounding boxes.

\begin{figure}[t!]
\centering
\begin{subfigure}{.32\linewidth}
  \centering
  \includegraphics[width=.9\linewidth]{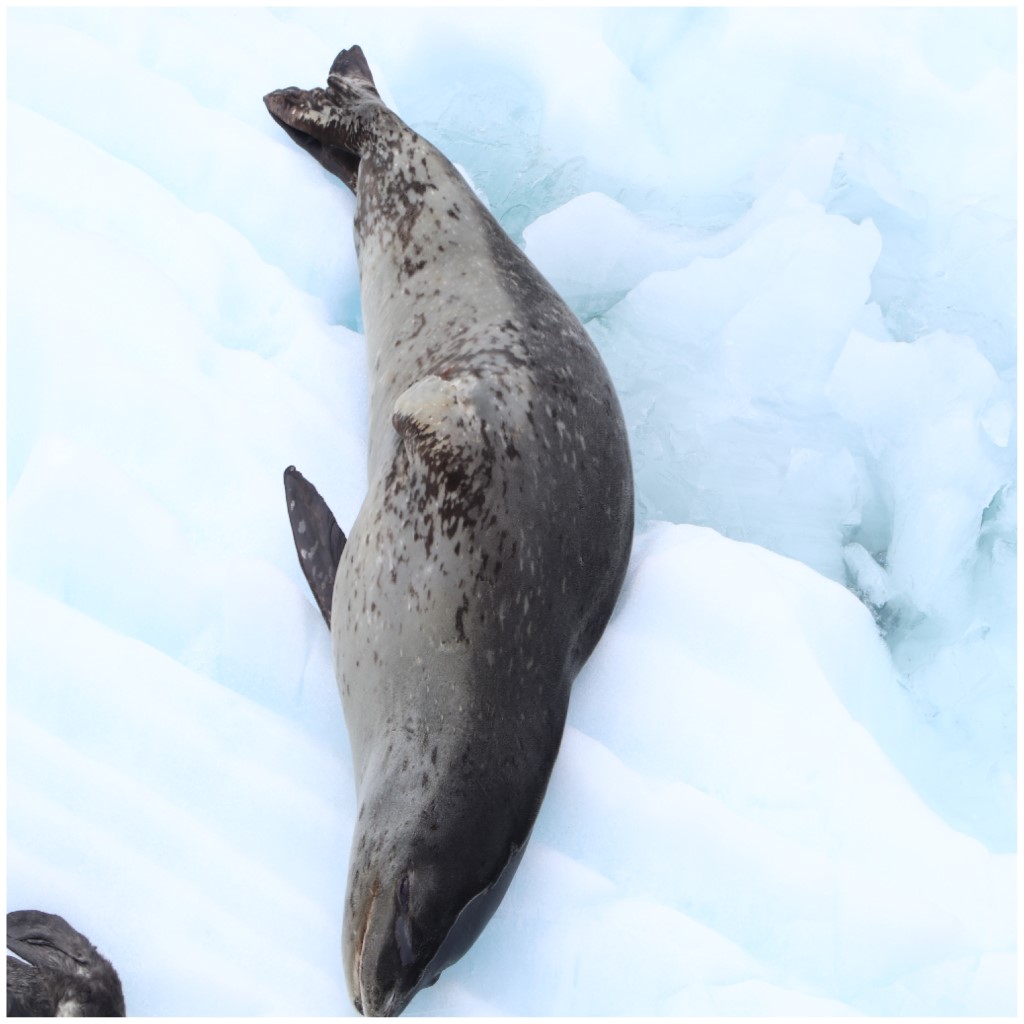}
  \caption{Original.}
  \label{fig:segmentation_example_seal}
\end{subfigure}%
\begin{subfigure}{.32\linewidth}
  \centering
  \includegraphics[width=.9\linewidth]{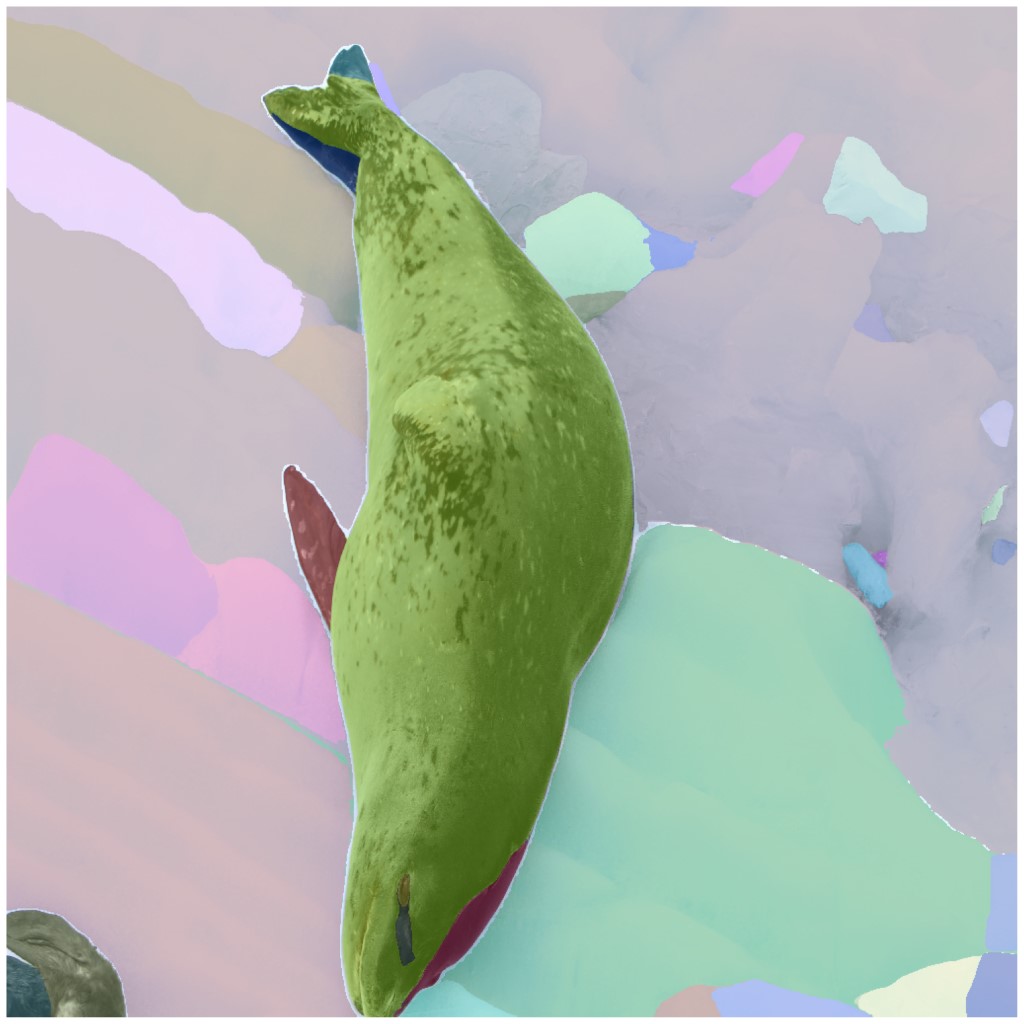}
  \caption{Automatic.}
  \label{fig:segmentation_example_seal_auto_sam}
\end{subfigure}
\begin{subfigure}{.32\linewidth}
  \centering
  \includegraphics[width=.9\linewidth]{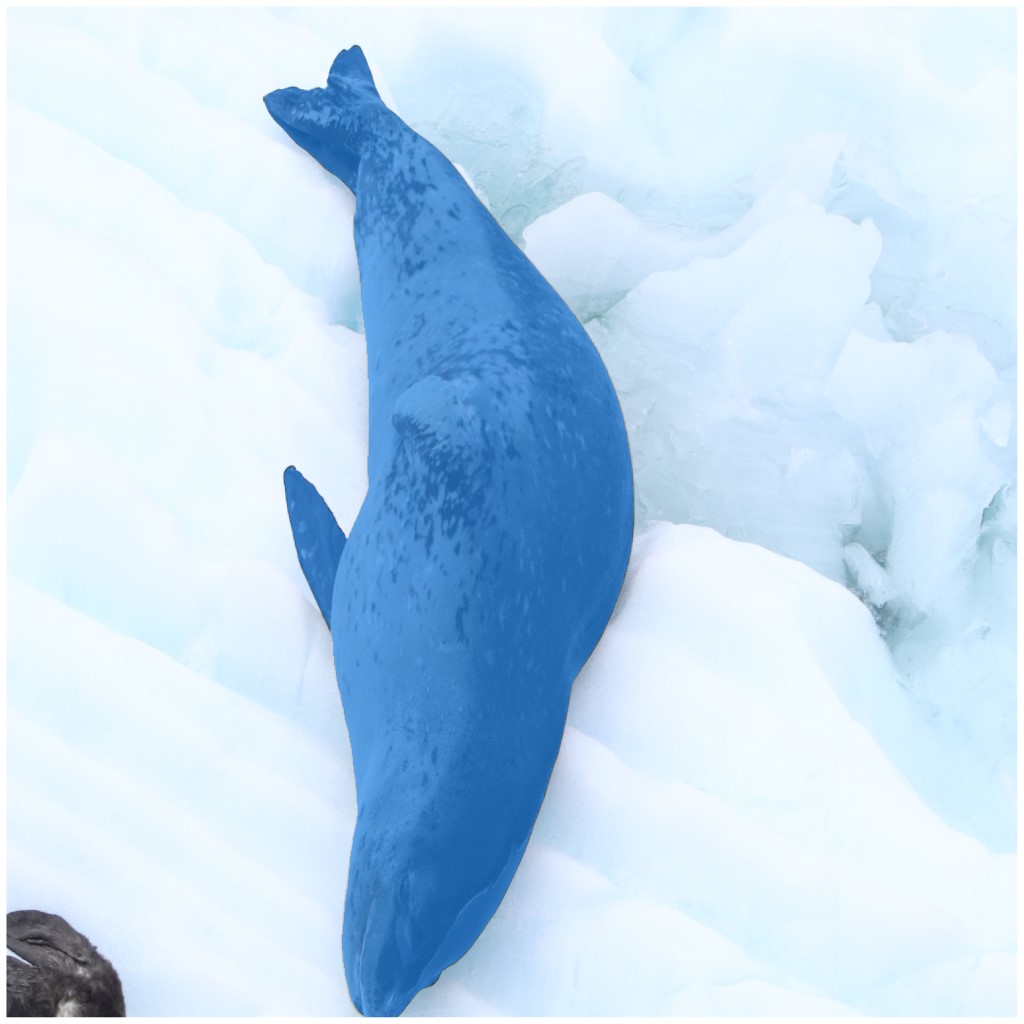}
  \caption{Bounding box.}
  \label{fig:segmentation_example_seal_box_sam}
\end{subfigure}
\caption{Seal image segmentation using SAM~\cite{segment_anything_kirillov2023segment}.}
\label{fig:segmentation_example}
\end{figure}

Conversely, despite its successes, the SAM ~\cite{segment_anything_kirillov2023segment} can sometimes produce small, disconnected segments within the seal imagery. It also falls short of real-time processing when employing a complex image encoder, thus lagging behind previous methods in terms of speed. SAM tends to over-segment when tasked with straightforward seal segmentation, capturing unnecessary details like skin patterns that aren't required for basic segmentation tasks. A proposed solution to mitigate this issue involves leveraging zero-shot transfer with a specified bounding box around the seal, guiding the model to generate a singular, comprehensive segmentation within this boundary. To evaluate this approach and its limitations, we experimented using both methods, with the results depicted in Fig. \ref{fig:segmentation_example}. As shown, \ref{fig:segmentation_example_seal_auto_sam} includes extraneous segments, whereas providing a bounding box in \ref{fig:segmentation_example_seal_box_sam} results in a singular, accurate segmentation of the seal, aligning with expectations. Although conclusions from a single image are not definitive, they align with anticipated outcomes based on the model's training to detect object subparts.

Altogether, seal segmentation presents significant challenges similar to those encountered in object detection, as outlined in Section \ref{sec:object_detection_discussion}. The primary obstacle is the need for prior research in this area, with limited studies addressing seal segmentation specifically. Compounding this issue is the critical shortage of annotated datasets, which are essential for training robust models. These datasets are particularly rare or challenging to compile, especially in the context of wildlife research. Although employing bounding boxes as input for the SAM represents a promising strategy for seal segmentation, the effective implementation of this approach still hinges on the availability of annotated datasets. These datasets, containing bounding boxes as ground truth, and are necessary to train object detection models. Consequently, the challenges previously identified for automatic detectors in section \ref{sec:object_detection} remain relevant and apply to the task of seal segmentation as well.

In summary, despite these challenges, SAM's demonstrated effectiveness and flexibility make it the preferred method for achieving our objectives of segmenting seals from conventional photographs, offering a promising avenue for advancing seal segmentation research.

\section{Vision transformer's popularity in the SOTA of image recognition. Potential application for seal recognition in marine environments}

The introduction of Vision Transformer (ViT,~\cite{vit_dosovitskiy2020image}) marked a significant milestone in the evolution of deep learning architectures. This innovation was heavily influenced by the unprecedented success of the Transformer model~\cite{transformer_vaswani2017attention} in the domain of natural language processing (NLP). The essence of ViT's innovation lies in its application of the standard Transformer architecture to the visual domain, which is implemented with minimal modifications. This approach underscores the versatility of the Transformer model, making an understanding of the original Transformer framework essential for fully appreciating the impact and functionality of ViT.

The Transformer model~\cite{transformer_vaswani2017attention}, revolutionized the field of machine translation. This success led to its rapid adoption across various NLP tasks. The profound impact of the Transformer model is evident when examining leading models across various NLP challenges. An outstanding example of its wide-ranging implications constitutes its dominance in the Multi-task Language Understanding (MMLU) benchmark~\cite{nlp_dataset_hendrycks2020measuring}, a comprehensive and challenging assessment covering 57 subjects from elementary to advanced professional levels, evaluating world knowledge and problem-solving abilities.

The Transformer architecture includes both an encoder and a decoder, but we will focus on the encoder because of its use in the ViT. It turns input tokens, $x_i$, into embeddings, $z_i$, of a specific size $D$. It adds positional information to these embeddings to keep track of the order of input, which is essential for understanding language and images. This positional information can be learned or set by rules to avoid looking ahead in the input. These enhanced embeddings then proceed to the multi-head attention layer, where the model calculates attention scores based on queries $Q$, keys $K$, and values $V$, as detailed in (\ref{eq:attention}) below. This part of the model decides how much focus to give each input part by comparing it against all other parts. It uses a mathematical operation (dot product) to calculate attention scores, which are turned into probabilities. These probabilities help select the essential parts of the input to produce an output relevant to the task being performed; attention is generally calculated as:
\begin{equation}\label{eq:attention}
    \text{Attention}(Q, K, V) = \text{softmax}(\frac{QK^T}{\sqrt{D}})V.
\end{equation}

A key factor behind the Transformer's success is its self-attention mechanism, which empowers the model with selective attention capabilities. This means that, following the self-attention process, the model has the potential to ``see'' all the input at once, understanding the context and the relationships between different parts of the input. Furthermore, the architecture presents significant capabilities for concurrent processing. Each attention head operates independently, allowing for simultaneous computation across heads. After computing attention, the representation of each token in the sequence becomes independent of others, enabling parallel processing across all tokens.

The ViT seeks to adhere closely to the original Transformer architecture. However, an unavoidable adaptation concerns the nature of the inputs; while the Transformer is designed for 1D sequences of tokens, ViT must handle 2D images. ViT divides an image into a sequence of $N$ patches, each of size $(P, P)$, to bridge this gap. These image patches, represented as $x_p \in \mathbb{R}^{N\times(P^2 C)}$ where $C$ denotes the number of image channels, are then mapped to a dimension of $N\times D$, aligning with the encoder's dimensionality $D$. Similar to BERT's approach~\cite{devlin2018bert}, a particular $x_{class}$ token is a prefix to this sequence. This token is designed to capture the overall representation of the image. The mapping of image patches to the model's dimension and their positional encoding within the image are described by (\ref{eq:vit_equation}) below, where $E$ performs the dimensional mapping, and $E_{pos}$ applies positional encoding. A classification head is attached to the output representation of the artificial $x_{class}$ token for classification tasks. Formally, the process is defined as:
\begin{equation}\label{eq:vit_equation}
    z_0 = [x_{class}; x^1_pE; x^2_pE; \cdots ; x^N_p E] + E_{pos} ,
\end{equation}
where $E \in \mathbb{R}^{(P^2C)\times D}$ and $E_{pos} \in \mathbb{R}^{(N+1)\times D}$.

\subsubsection{Vision Transformer; the popular option}

Since the advent of AlexNet~\cite{alex_net_krizhevsky2012imagenet}, Convolutional Neural Networks (CNNs) have been the gold standard in computer vision, dominating the field with their state-of-the-art performance. However, the emergence of Vision Transformers (ViTs) has introduced a paradigm shift. Unlike CNNs, which rely on local receptive fields to process images incrementally, ViTs leverage self-attention mechanisms to analyze entire images holistically. Achieving a similar level of global attention with CNNs would require stacking multiple layers, a process that significantly restrains efficiency compared to the streamlined performance of transformers. This global approach allows ViTs to capture long-range dependencies within the data, a feature especially beneficial for complex recognition tasks where understanding context and spatial relationships is paramount.

ViTs have challenged the supremacy of CNNs and showcased exceptional scalability and computational efficiency. As the size of the model scales, ViTs tend to exhibit improved performance, making them exceptionally well-suited for handling large datasets. This scalability, combined with the inherent parallelizability of self-attention mechanisms—more so than convolution operations—positions ViTs as a more computationally efficient choice for extensive tasks~\cite{alex_net_krizhevsky2012imagenet}.

Furthermore, ViTs have demonstrated significant expertise in transfer learning scenarios. They excel at adapting knowledge from extensive datasets to specific tasks with limited data, maintaining high-performance levels. This versatility is invaluable in wildlife monitoring, where obtaining labeled data can be challenging. The availability of pre-trained ViT models of various sizes (82M, 307M, and 632M) further simplifies fine-tuning for specific applications. Collectively, these attributes underscore the growing popularity and potential of ViTs to redefine the landscape of image recognition and beyond, marking a significant milestone in the evolution of computer vision technologies.

\subsection{Potential of ViT for recognizing/identifying seals in surface marine environments}

Regarding the potential of Vision Transformers (ViTs) for recognizing and identifying seals in surface marine environments, their capabilities extend across several key areas:

Fine-grained Classification: ViTs excel in fine-grained classification tasks~\cite{he2022transfg}, making them highly effective at distinguishing between closely related seal species or identifying individual seals within a species. This is due to their ability to focus on fine details across the entire image, capturing subtle differences that might elude other models.

Robustness to Environmental Variability: The inherent global perspective of ViTs equips them with a significant advantage in dealing with the environmental variability typical of marine settings, such as changes in lighting, background, and weather conditions~\cite{paul2022vision}. This robustness enhances their reliability and accuracy in real-world applications.

Transfer Learning for Rare Species: ViTs demonstrate exceptional transfer learning capabilities, which can be particularly valuable for recognizing rare or endangered seal species~\cite{li2021benchmarking}. By fine-tuning pre-trained models on more common species or related tasks, researchers can develop effective recognition systems even with scarce training data for specific rare species~\cite{detection_cetacean_conventional_patton2023deep}.

High-Quality Image Embeddings: Utilizing a methodology similar to the image encoder for SAM~\cite{segment_anything_kirillov2023segment} or for ImageBind~\cite{girdhar2023imagebind}, a ViT can generate high-quality image embeddings. These embeddings can serve as a robust foundation for seal recognition tasks, offering a sophisticated approach to understanding and classifying marine wildlife imagery.

In summary, ViTs are popular because they efficiently process and understand complex image data at once through the attention mechanism~\cite{vit_dosovitskiy2020image}, making them particularly promising for specialized applications like recognizing seals in marine environments.

\section{Challenges in seal recognition from the perspectives of Computer Vision and Machine Learning. Impact over the development and performance of automated detection systems}

From the perspectives of Computer Vision (CV) and Machine Learning (ML), several primary challenges are associated with the recognition and identification of seals, impacting both the development and performance of automated detection systems. These challenges include:

\textbf{Data:} High-quality, annotated datasets of seals are crucial for training effective ML models. However, collecting and annotating such datasets is labor-intensive and challenging, particularly for less common species or those living in remote locations. To alleviate this issue, data augmentation techniques using Generative Adversarial Networks (GANs)~\cite{gan_goodfellow2014generative} or Diffusion Models~\cite{sohl2015deep} are used to increase the size of the dataset for training time.

\begin{figure}[h!]
    \centering
    \begin{subfigure}{.45\linewidth}
    \centering
        \includegraphics[width=.6\linewidth]{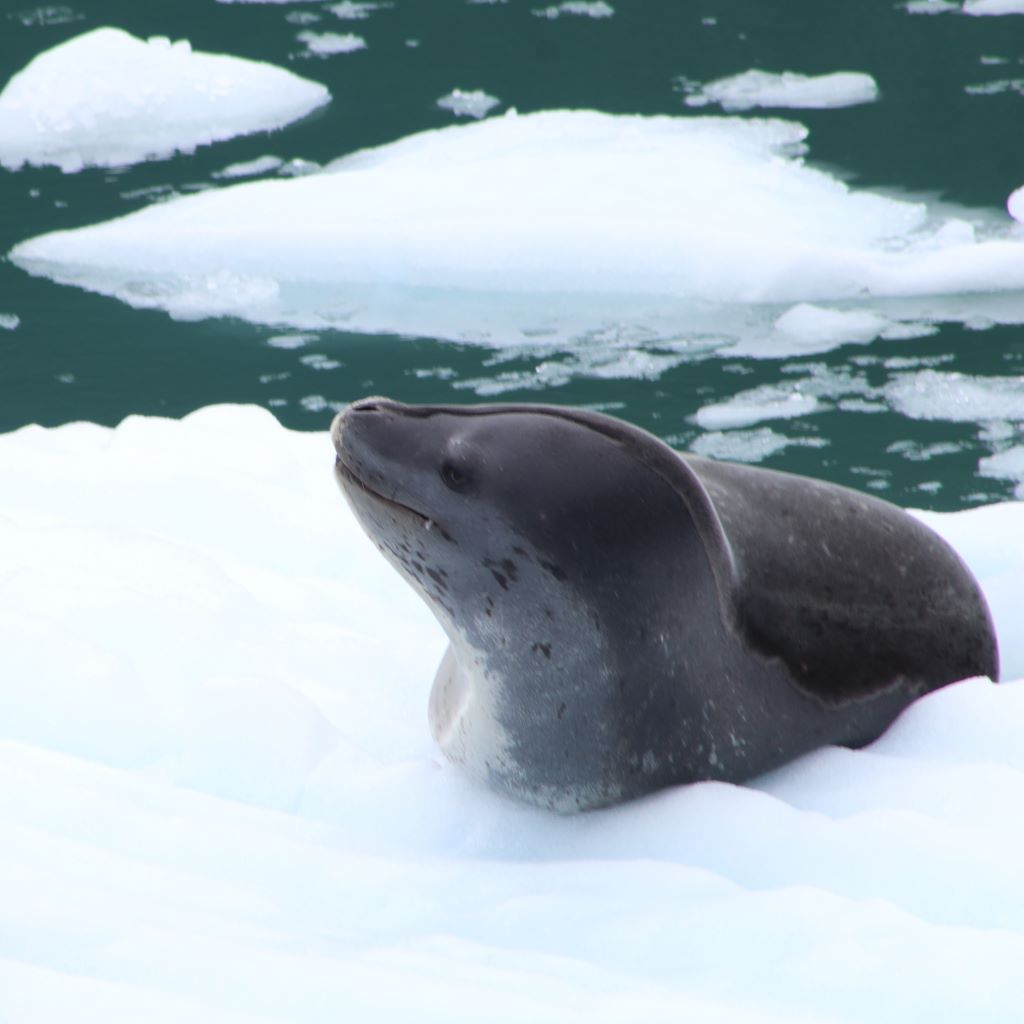}
        \caption{Leopard Seal left side of face.}
        \label{fig:seal_pose_left}
    \end{subfigure}
    \begin{subfigure}{.45\linewidth}
    \centering
        \includegraphics[width=.6\linewidth]{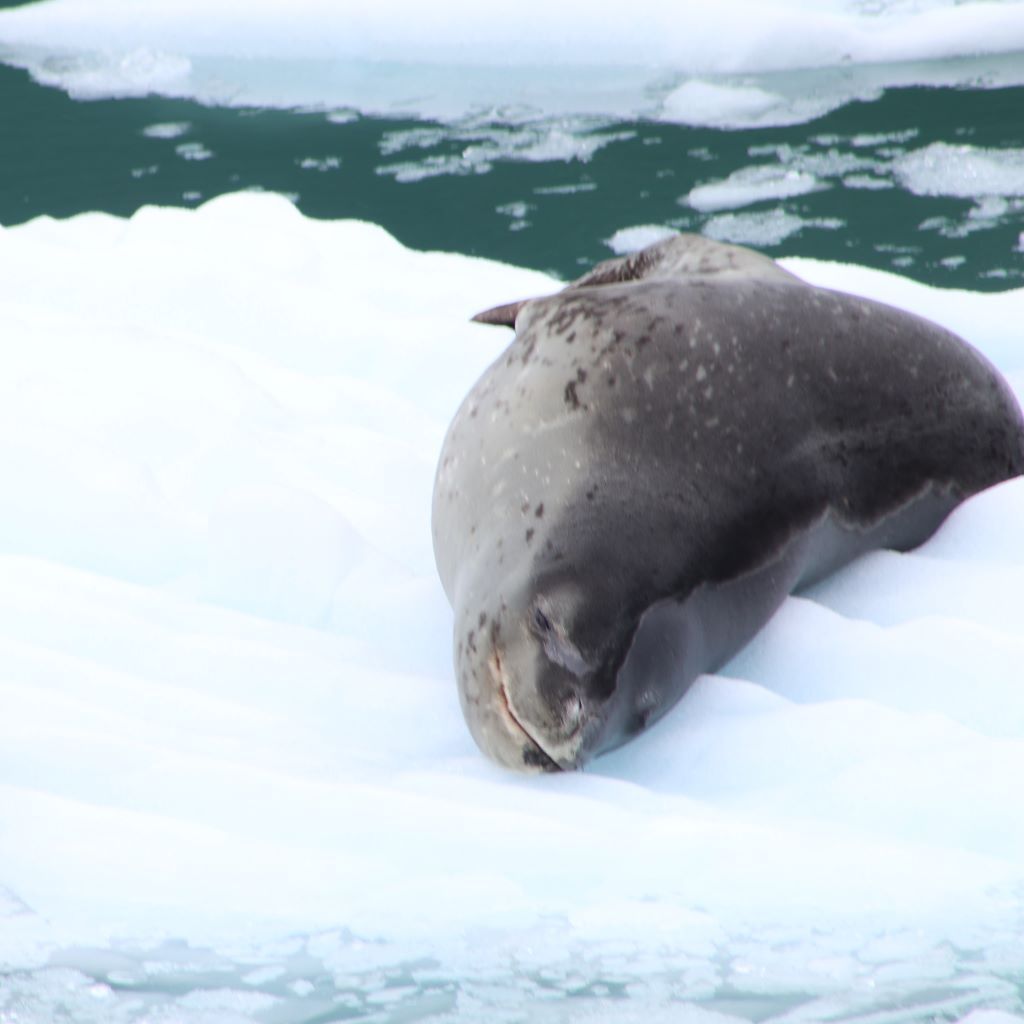}
        \caption{Leopard seal right side of full body.}
        \label{fig:seal_pose_right}
    \end{subfigure}
    \caption{Samples of leopard seals on different body sides.}
    \label{fig:leopard_seal}
\end{figure}

\textbf{Pose Variation:} Seals can adopt various poses. The pose of the seal is essential for individual recognition; if we have the right side of a seal, we don't want to try to match it with the left side of a seal. A potential solution is to detect the pose of the seal and combine the pose and the features of the seal; a possible approach is using learnable embedding that represents the pose of the seal and applies it to the seal itself. Fig. \ref{fig:leopard_seal} shows various angles of the same seal, which appear dissimilar due to the images capturing different sides of its body.

\begin{figure}[h!]
    \centering
    \begin{subfigure}[t]{.32\linewidth}
    \centering
        \includegraphics[width=.88\linewidth]{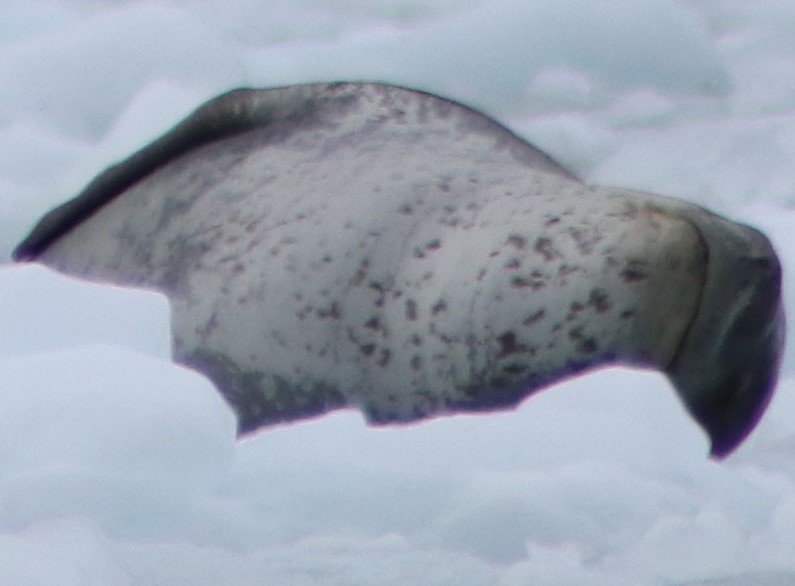}
        \caption{Leopard seal occluded by ice.}
        \label{fig:seal_occlusion}
    \end{subfigure}
    \begin{subfigure}[t]{.32\linewidth}
    \centering
        \includegraphics[width=.88\linewidth]{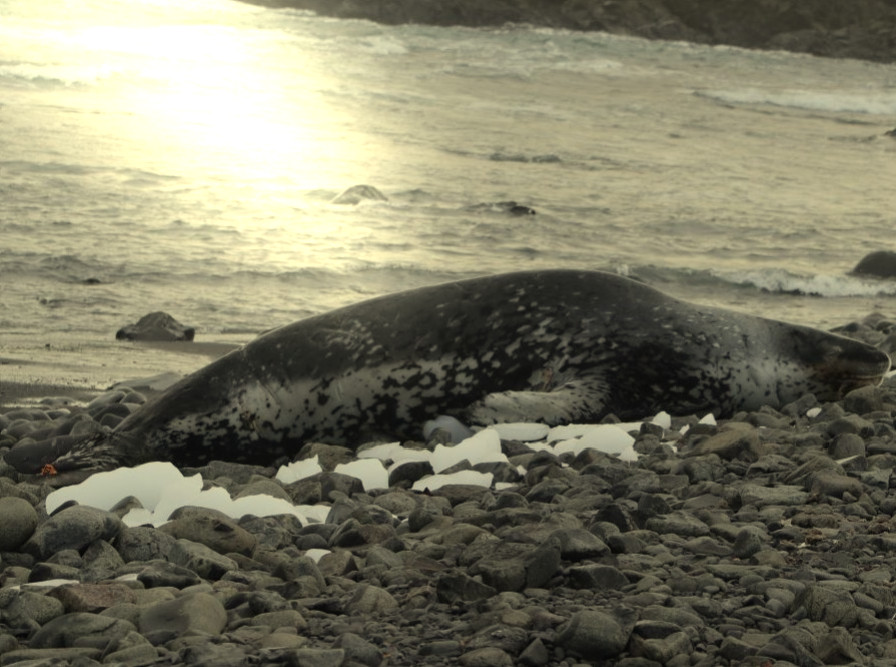}
        \caption{Leopard seal image affected by sunlight.}
        \label{fig:seal_weather}
    \end{subfigure}
    \begin{subfigure}[t]{.32\linewidth}
    \centering
        \includegraphics[width=.88\linewidth]{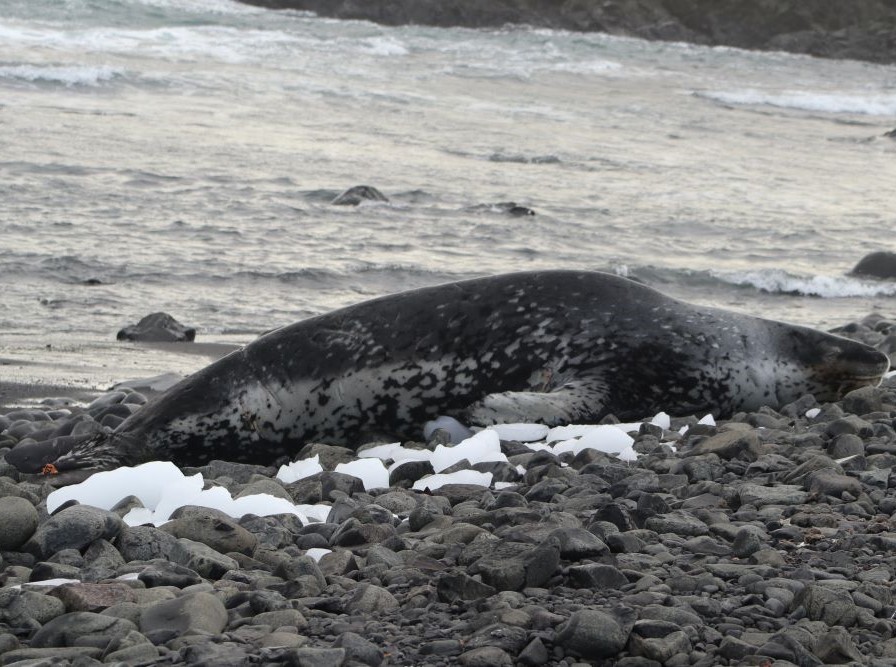}
        \caption{Leopard seal hauling-out onshore in Antarctica.}
        \label{fig:seal_background}
    \end{subfigure}
    \caption{Samples that represent different data challenges.}
    \label{fig:data_challenges}
\end{figure}

\textbf{Background and occlusion:} The natural habitats of seals vary from glacial habitats with icebergs acting as the main haul-out substrates for the animals to sandy or rocky beaches. Also, they may be partially occluded by objects in their environment or by other seals. These conditions push the automatic algorithm to extract the correct features of the seal and ignore the background or object in front of the seal. An approach to mitigate this issue is to segment the seal, removing all the background noise and the occluded objects. Fig. \ref{fig:seal_background} illustrates a seal in a non-ice background, while Fig. \ref{fig:seal_occlusion} depicts an example of a seal occluded by an object.

\textbf{Weather conditions:} Lighting conditions can dramatically affect the appearance of seals in images, particularly for those captured in the wild. Variations in weather, time of day, and water reflections can alter how seals are visualized, affecting detection consistency. To reduce this issue, data augmentation techniques can augment the dataset by applying different filters to the image, like blurring and sepia. Fig. \ref{fig:seal_weather} shows an example of a seal where the colors are different due to sunlight.

\begin{figure}[h!]
    \centering
    \includegraphics[width=.3\linewidth]{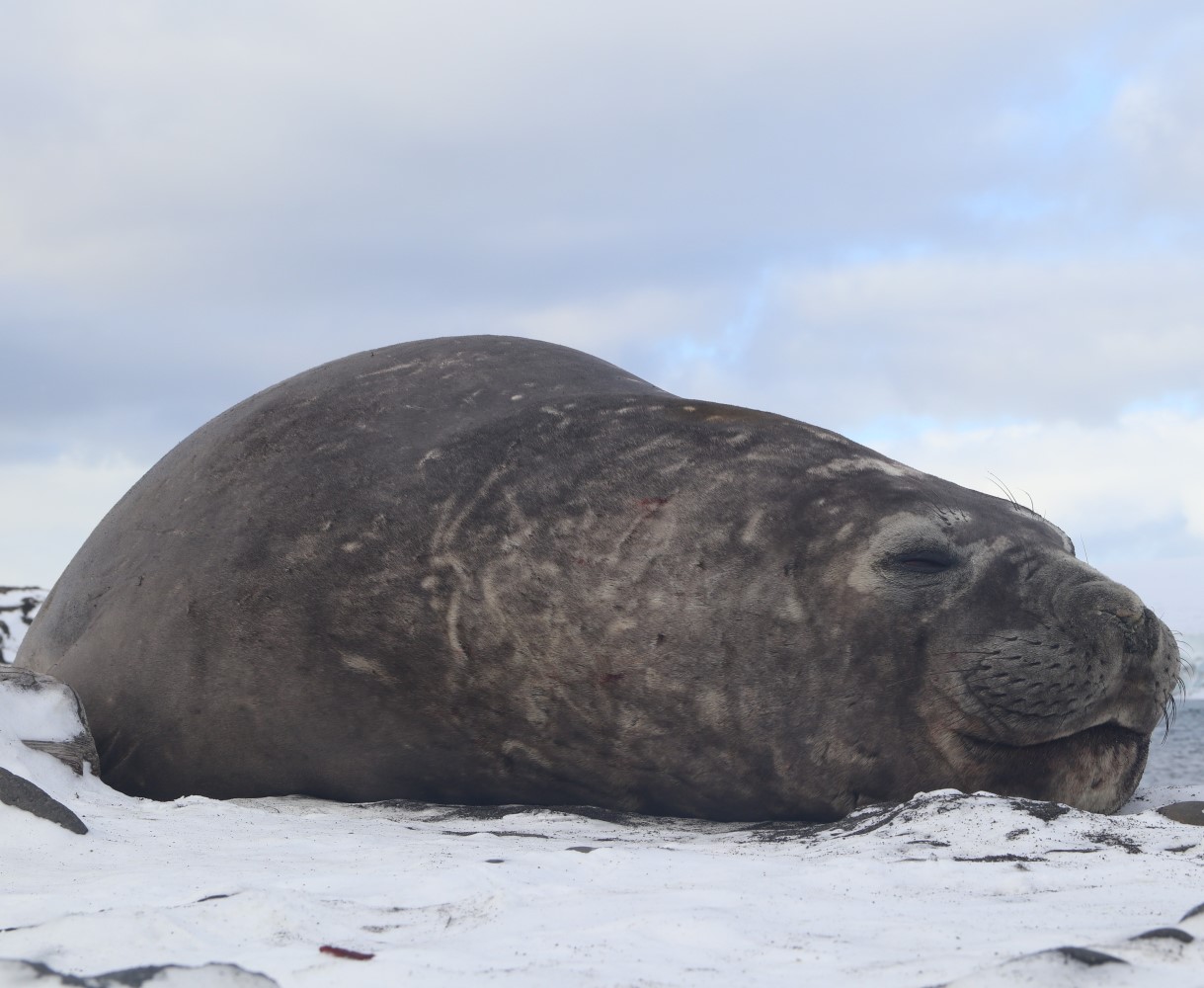}
    \caption{Southern elephant seal (\emph{Mirounga leonina}) a different species than our model species leopard seals.}
    \label{fig:seal_variability}
\end{figure}

\textbf{Variability:} Seals can vary significantly in color, size, and shape due to age, species, molting stage, and environmental factors. To overcome this issue, a good dataset with enough samples of each one will be ideal. Still, if we don't possess enough samples from all the conditions, we can try an approach similar to~\cite{detection_cetacean_conventional_patton2023deep} where we add an additional head to the model. Hence, it can produce species besides the identification. In this way, the idea of the model is that it can transfer learning from similar species to the ones that contain just a few images. Fig. \ref{fig:seal_variability} illustrates an example of a different seal species (a southern elephant seal, \emph{Mirouga Leonina}) compared to our model species used in Figs. \ref{fig:leopard_seal} and \ref{fig:data_challenges} with specific features in the body and the face that differentiate them.

\section{Conclusion}

Our investigation highlights the transformative potential of machine learning, particularly vision transformers, for seal recognition using conventional photography. Although traditional methods of seal monitoring present challenges in scalability and adaptability, machine learning emerges as a robust solution, effectively bridging these gaps. Despite their precision, manual detection methods are limited by their lack of scalability. In contrast, automated approaches like the YOLO framework offer scalability and potential for real-time application, yet their efficacy across diverse environmental settings warrants further empirical exploration.

One of the primary obstacles in this domain is the variability inherent in conventional photographic data, influenced by factors like image quality and environmental conditions. Despite these challenges, the advent of machine learning and, specifically, vision transformers, signals a promising avenue for advancing seal monitoring processes. These technologies have the potential to significantly enhance the precision and efficiency of species identification efforts, providing a pathway to more effective, non-intrusive wildlife monitoring.

This research contributes to the broader understanding of applying image processing techniques within wildlife monitoring and conservation science, suggesting that integrating CV and ML could revolutionize environmental conservation practices. Through the lens of ML, we envision a future where monitoring and protecting seal species in their natural habitats is more effective and less intrusive.

\section*{Acknowledgments}
Part of this work was funded by the National Science Foundation under grants CNS-2210091, OPP-2146068, CHE-1905043, and CNS-2136961.

% \IEEEtriggeratref{36}
\bibliography{refs}

\begin{thebibliography}{10}
\providecommand{\url}[1]{#1}
\csname url@samestyle\endcsname
\providecommand{\newblock}{\relax}
\providecommand{\bibinfo}[2]{#2}
\providecommand{\BIBentrySTDinterwordspacing}{\spaceskip=0pt\relax}
\providecommand{\BIBentryALTinterwordstretchfactor}{4}
\providecommand{\BIBentryALTinterwordspacing}{\spaceskip=\fontdimen2\font plus
\BIBentryALTinterwordstretchfactor\fontdimen3\font minus \fontdimen4\font\relax}
\providecommand{\BIBforeignlanguage}[2]{{%
\expandafter\ifx\csname l@#1\endcsname\relax
\typeout{** WARNING: IEEEtran.bst: No hyphenation pattern has been}%
\typeout{** loaded for the language `#1'. Using the pattern for}%
\typeout{** the default language instead.}%
\else
\language=\csname l@#1\endcsname
\fi
#2}}
\providecommand{\BIBdecl}{\relax}
\BIBdecl

\bibitem{pritt2017satellite}
M.~Pritt and G.~Chern, ``Satellite image classification with deep learning,'' in \emph{2017 IEEE applied imagery pattern recognition workshop (AIPR)}.\hskip 1em plus 0.5em minus 0.4em\relax IEEE, 2017, pp. 1--7.

\bibitem{detection_seal_aerial_salberg2015detection}
A.-B. Salberg, ``Detection of seals in remote sensing images using features extracted from deep convolutional neural networks,'' in \emph{2015 IEEE International Geoscience and Remote Sensing Symposium (IGARSS)}.\hskip 1em plus 0.5em minus 0.4em\relax IEEE, 2015, pp. 1893--1896.

\bibitem{detection_seal_satellite_gonccalves2020sealnet}
B.~C. Gon{\c{c}}alves, B.~Spitzbart, and H.~J. Lynch, ``Sealnet: A fully-automated pack-ice seal detection pipeline for sub-meter satellite imagery,'' \emph{Remote Sensing of Environment}, vol. 239, p. 111617, 2020.

\bibitem{detection_seal_satellite_gonccalves2022sealnet2}
B.~C. Gon{\c{c}}alves, M.~Wethington, and H.~J. Lynch, ``Sealnet 2.0: Human-level fully-automated pack-ice seal detection in very-high-resolution satellite imagery with cnn model ensembles,'' \emph{Remote Sensing}, vol.~14, no.~22, p. 5655, 2022.

\bibitem{detection_primates_conventional_manual_deb2018face}
D.~Deb, S.~Wiper, S.~Gong, Y.~Shi, C.~Tymoszek, A.~Fletcher, and A.~K. Jain, ``Face recognition: Primates in the wild,'' in \emph{2018 IEEE 9th International Conference on Biometrics Theory, Applications and Systems (BTAS)}.\hskip 1em plus 0.5em minus 0.4em\relax IEEE, 2018, pp. 1--10.

\bibitem{detection_lemur_conventional_manual_crouse2017lemurfaceid}
D.~Crouse, R.~L. Jacobs, Z.~Richardson, S.~Klum, A.~Jain, A.~L. Baden, and S.~R. Tecot, ``Lemurfaceid: A face recognition system to facilitate individual identification of lemurs,'' \emph{Bmc Zoology}, vol.~2, no.~1, pp. 1--14, 2017.

\bibitem{detection_cetacean_conventional_patton2023deep}
P.~T. Patton, T.~Cheeseman, K.~Abe, T.~Yamaguchi, W.~Reade, K.~Southerland, A.~Howard, E.~M. Oleson, J.~B. Allen, E.~Ashe \emph{et~al.}, ``A deep learning approach to photo--identification demonstrates high performance on two dozen cetacean species,'' \emph{Methods in ecology and evolution}, vol.~14, no.~10, pp. 2611--2625, 2023.

\bibitem{detection_seal_conventional_manual_birenbaum2022sealnet}
Z.~Birenbaum, H.~Do, L.~Horstmyer, H.~Orff, K.~Ingram, and A.~Ay, ``Sealnet: Facial recognition software for ecological studies of harbor seals,'' \emph{Ecology and Evolution}, vol.~12, no.~5, p. e8851, 2022.

\bibitem{cnn_lecun1989backpropagation}
Y.~LeCun, B.~Boser, J.~S. Denker, D.~Henderson, R.~E. Howard, W.~Hubbard, and L.~D. Jackel, ``Backpropagation applied to handwritten zip code recognition,'' \emph{Neural computation}, vol.~1, no.~4, pp. 541--551, 1989.

\bibitem{yolov1_redmon2016you}
J.~Redmon, S.~Divvala, R.~Girshick, and A.~Farhadi, ``You only look once: Unified, real-time object detection,'' in \emph{Proceedings of the IEEE conference on computer vision and pattern recognition}, 2016, pp. 779--788.

\bibitem{detic_zhou2022detecting}
X.~Zhou, R.~Girdhar, A.~Joulin, P.~Kr{\"a}henb{\"u}hl, and I.~Misra, ``Detecting twenty-thousand classes using image-level supervision,'' in \emph{European Conference on Computer Vision}.\hskip 1em plus 0.5em minus 0.4em\relax Springer, 2022, pp. 350--368.

\bibitem{face_recognition_jain2011handbook}
A.~K. Jain and S.~Z. Li, \emph{Handbook of face recognition}.\hskip 1em plus 0.5em minus 0.4em\relax Springer, 2011, vol.~1.

\bibitem{arcface_deng2019arcface}
J.~Deng, J.~Guo, N.~Xue, and S.~Zafeiriou, ``Arcface: Additive angular margin loss for deep face recognition,'' in \emph{Proceedings of the IEEE/CVF conference on computer vision and pattern recognition}, 2019, pp. 4690--4699.

\bibitem{arcface_deng2020sub}
J.~Deng, J.~Guo, T.~Liu, M.~Gong, and S.~Zafeiriou, ``Sub-center arcface: Boosting face recognition by large-scale noisy web faces,'' in \emph{Computer Vision--ECCV 2020: 16th European Conference, Glasgow, UK, August 23--28, 2020, Proceedings, Part XI 16}.\hskip 1em plus 0.5em minus 0.4em\relax Springer, 2020, pp. 741--757.

\bibitem{seal_segmentation_zhelezniakov2015segmentation}
A.~Zhelezniakov, T.~Eerola, M.~Koivuniemi, M.~Auttila, R.~Lev{\"a}nen, M.~Niemi, M.~Kunnasranta, and H.~K{\"a}lvi{\"a}inen, ``Segmentation of saimaa ringed seals for identification purposes,'' in \emph{Advances in Visual Computing: 11th International Symposium, ISVC 2015, Las Vegas, NV, USA, December 14-16, 2015, Proceedings, Part II 11}.\hskip 1em plus 0.5em minus 0.4em\relax Springer, 2015, pp. 227--236.

\bibitem{unsupervised_semantic_segmentation_arbelaez2009contours}
P.~Arbelaez, M.~Maire, C.~Fowlkes, and J.~Malik, ``From contours to regions: An empirical evaluation,'' in \emph{2009 IEEE Conference on Computer Vision and Pattern Recognition}.\hskip 1em plus 0.5em minus 0.4em\relax IEEE, 2009, pp. 2294--2301.

\bibitem{unsupervised_semantic_segmentation_arbelaez2010contour}
------, ``Contour detection and hierarchical image segmentation,'' \emph{IEEE transactions on pattern analysis and machine intelligence}, vol.~33, no.~5, pp. 898--916, 2010.

\bibitem{ojansivu2008blur}
V.~Ojansivu and J.~Heikkil{\"a}, ``Blur insensitive texture classification using local phase quantization,'' in \emph{Image and Signal Processing: 3rd International Conference, ICISP 2008. Cherbourg-Octeville, France, July 1-3, 2008. Proceedings 3}.\hskip 1em plus 0.5em minus 0.4em\relax Springer, 2008, pp. 236--243.

\bibitem{segmentation_animals_water_pollicelli2020roi}
D.~Pollicelli, M.~Coscarella, and C.~Delrieux, ``Roi detection and segmentation algorithms for marine mammals photo-identification,'' \emph{Ecological Informatics}, vol.~56, p. 101038, 2020.

\bibitem{segmentation_animalli2021marine}
L.~Li, B.~Dong, E.~Rigall, T.~Zhou, J.~Dong, and G.~Chen, ``Marine animal segmentation,'' \emph{IEEE Transactions on Circuits and Systems for Video Technology}, vol.~32, no.~4, pp. 2303--2314, 2021.

\bibitem{segment_anything_kirillov2023segment}
A.~Kirillov, E.~Mintun, N.~Ravi, H.~Mao, C.~Rolland, L.~Gustafson, T.~Xiao, S.~Whitehead, A.~C. Berg, W.-Y. Lo \emph{et~al.}, ``Segment anything,'' \emph{arXiv preprint arXiv:2304.02643}, 2023.

\bibitem{mae_he2022masked}
K.~He, X.~Chen, S.~Xie, Y.~Li, P.~Doll{\'a}r, and R.~Girshick, ``Masked autoencoders are scalable vision learners,'' in \emph{Proceedings of the IEEE/CVF conference on computer vision and pattern recognition}, 2022, pp. 16\,000--16\,009.

\bibitem{vit_dosovitskiy2020image}
A.~Dosovitskiy, L.~Beyer, A.~Kolesnikov, D.~Weissenborn, X.~Zhai, T.~Unterthiner, M.~Dehghani, M.~Minderer, G.~Heigold, S.~Gelly \emph{et~al.}, ``An image is worth 16x16 words: Transformers for image recognition at scale,'' \emph{arXiv preprint arXiv:2010.11929}, 2020.

\bibitem{vitdec_li2022exploring}
Y.~Li, H.~Mao, R.~Girshick, and K.~He, ``Exploring plain vision transformer backbones for object detection,'' in \emph{European Conference on Computer Vision}.\hskip 1em plus 0.5em minus 0.4em\relax Springer, 2022, pp. 280--296.

\bibitem{coco_lin2014microsoft}
T.-Y. Lin, M.~Maire, S.~Belongie, J.~Hays, P.~Perona, D.~Ramanan, P.~Doll{\'a}r, and C.~L. Zitnick, ``Microsoft coco: Common objects in context,'' in \emph{Computer Vision--ECCV 2014: 13th European Conference, Zurich, Switzerland, September 6-12, 2014, Proceedings, Part V 13}.\hskip 1em plus 0.5em minus 0.4em\relax Springer, 2014, pp. 740--755.

\bibitem{lvis_gupta2019lvis}
A.~Gupta, P.~Dollar, and R.~Girshick, ``Lvis: A dataset for large vocabulary instance segmentation,'' in \emph{Proceedings of the IEEE/CVF conference on computer vision and pattern recognition}, 2019, pp. 5356--5364.

\bibitem{transformer_vaswani2017attention}
A.~Vaswani, N.~Shazeer, N.~Parmar, J.~Uszkoreit, L.~Jones, A.~N. Gomez, {\L}.~Kaiser, and I.~Polosukhin, ``Attention is all you need,'' \emph{Advances in neural information processing systems}, vol.~30, 2017.

\bibitem{nlp_dataset_hendrycks2020measuring}
D.~Hendrycks, C.~Burns, S.~Basart, A.~Zou, M.~Mazeika, D.~Song, and J.~Steinhardt, ``Measuring massive multitask language understanding,'' \emph{arXiv preprint arXiv:2009.03300}, 2020.

\bibitem{devlin2018bert}
J.~Devlin, M.-W. Chang, K.~Lee, and K.~Toutanova, ``Bert: Pre-training of deep bidirectional transformers for language understanding,'' \emph{arXiv preprint arXiv:1810.04805}, 2018.

\bibitem{alex_net_krizhevsky2012imagenet}
A.~Krizhevsky, I.~Sutskever, and G.~E. Hinton, ``Imagenet classification with deep convolutional neural networks,'' \emph{Advances in neural information processing systems}, vol.~25, 2012.

\bibitem{he2022transfg}
J.~He, J.-N. Chen, S.~Liu, A.~Kortylewski, C.~Yang, Y.~Bai, and C.~Wang, ``Transfg: A transformer architecture for fine-grained recognition,'' in \emph{Proceedings of the AAAI conference on artificial intelligence}, vol.~36, no.~1, 2022, pp. 852--860.

\bibitem{paul2022vision}
S.~Paul and P.-Y. Chen, ``Vision transformers are robust learners,'' in \emph{Proceedings of the AAAI conference on Artificial Intelligence}, vol.~36, no.~2, 2022, pp. 2071--2081.

\bibitem{li2021benchmarking}
Y.~Li, S.~Xie, X.~Chen, P.~Dollar, K.~He, and R.~Girshick, ``Benchmarking detection transfer learning with vision transformers,'' \emph{arXiv preprint arXiv:2111.11429}, 2021.

\bibitem{girdhar2023imagebind}
R.~Girdhar, A.~El-Nouby, Z.~Liu, M.~Singh, K.~V. Alwala, A.~Joulin, and I.~Misra, ``Imagebind: One embedding space to bind them all,'' in \emph{Proceedings of the IEEE/CVF Conference on Computer Vision and Pattern Recognition}, 2023, pp. 15\,180--15\,190.

\bibitem{gan_goodfellow2014generative}
I.~Goodfellow, J.~Pouget-Abadie, M.~Mirza, B.~Xu, D.~Warde-Farley, S.~Ozair, A.~Courville, and Y.~Bengio, ``Generative adversarial nets,'' \emph{Advances in neural information processing systems}, vol.~27, 2014.

\bibitem{sohl2015deep}
J.~Sohl-Dickstein, E.~Weiss, N.~Maheswaranathan, and S.~Ganguli, ``Deep unsupervised learning using nonequilibrium thermodynamics,'' in \emph{International conference on machine learning}.\hskip 1em plus 0.5em minus 0.4em\relax PMLR, 2015, pp. 2256--2265.

\end{thebibliography}

\end{document}